\pdfoutput=1

\documentclass[11pt]{article}

\usepackage[]{acl}

\usepackage{times}
\usepackage{latexsym}

\usepackage[T1]{fontenc}

\usepackage[utf8]{inputenc}

\usepackage{microtype}

\usepackage{inconsolata}

\usepackage{booktabs}       
\usepackage{float}  
 \usepackage{graphicx}
\usepackage{algorithm}
\usepackage{multirow}

\newcommand{\namoperator}[1]{\mathit{#1}}

\title{Deferred NAM: Low-latency Top-K Context Injection via Deferred Context Encoding for Non-Streaming ASR}

\author{Zelin Wu $\qquad$ Gan Song $\qquad$ Christopher Li $\qquad$ Pat Rondon $\qquad$ Zhong Meng \\
{\bf Xavier Velez} $\qquad$ {\bf Weiran Wang} $\qquad$ {\bf Diamantino Caseiro} $\qquad$ {\bf Golan Pundak} \\
{\bf Tsendsuren Munkhdalai} $\qquad$ {\bf Angad Chandorkar} $\qquad$ {\bf Rohit Prabhavalkar} \\
  Google LLC, USA \\
  \texttt{\{zelinwu, gansong, chriswli, rondon, zhongmeng, xavvelez\}@google.com}}

\begin{document}
\maketitle
\begin{abstract}
Contextual biasing enables speech recognizers to transcribe important phrases in the speaker’s context, such as contact names, even if they are rare in, or absent from, the training data. Attention-based biasing is a leading approach which allows for full end-to-end cotraining of the recognizer and biasing system and requires no separate inference-time components. Such biasers typically consist of a context encoder; followed by a context filter which narrows down the context to apply, improving per-step inference time; and, finally, context application via cross attention. Though much work has gone into optimizing per-frame performance, the context encoder is at least as important: recognition cannot begin before context encoding ends. Here, we
show the lightweight phrase selection pass can be moved before context encoding, resulting in a speedup of up to 16.1 times and enabling biasing to scale to 20K phrases with a maximum pre-decoding delay under 33ms. With the addition of phrase- and wordpiece-level cross-entropy losses, our technique also achieves up to a 37.5\% relative WER reduction over the baseline without the losses and lightweight phrase selection pass.
\end{abstract}

\section{Introduction}
\label{sec:introduction}

Automatic speech recognition (ASR) applications often succeed or fail based on their ability to recognize words that are relevant in context, but may not be common, or even present, in the training data.
For example, an assistant user may speak contact names from another language or titles of media entities which were released after the ASR system was trained, or the speaker may use domain-specific jargon, like legalese or medical terms, which are not common in the more-typical speech used for training.
ASR contextual biasing~\cite{Hall2015-ja,Aleksic2015-ob} aims to account for this domain shift between training and inference.

Attention-based biasing~\cite{Pundak2018-fj}, in which the context is encoded into dense embeddings attended to during recognition, is one of the leading approaches for contextualizing end-to-end (E2E) ASR systems.
As a fully-end-to-end method, it does not require separate biasing components which must be separately trained and whose integration with the core ASR system must be optimized.
However, recognition cannot begin until the inference-time context has been encoded, and this context may consist of tens of thousands of items which may not be cached, for privacy or system design reasons or, more simply, because the context may change at any point up until the beginning of recognition.
Thus, the context encoder must be able to efficiently handle very large contexts, as delays in context encoding translate directly into user-visible delays in ASR transcription. 

In this work, we optimize context encoding by splitting it into two passes.
We make the simplifying assumption that our ASR system is non-causal and can access the entire audio input;
we argue that this choice is not overly restrictive, as two-pass ASR systems combining a causal ASR system to produce streaming results with a non-causal ASR system for producing the final result have become common, e.g., as in~\cite{Narayanan2021-vi}.
With the speaker’s entire audio available before context encoding begins, we can split context encoding into two phases.
First, we encode all contextual phrases with an extremely lightweight encoder, and use the resulting encodings to determine the $k$ phrases that are most likely to occur in the audio.
We embed only the $k$ most-likely phrases with a more powerful --- but more expensive --- encoder, and use the output of this ``deferred'' encoder for biasing.

\section{Related Work}

Conventional ASR contextualization relies on discrete contextualization components like (class-based) language models (LMs)~\cite{Vasserman2016-el}, combined with the base ASR system through on-the-fly LM rescoring~\cite{Aleksic2015-ob,Hall2015-ja,McGraw2016-kv}, shallow fusion~\cite{Williams2018-ax,Zhao2019-ay}, or lattice rewriting~\cite{Serrino2019-yf} and contextual spelling correctors~\cite{Wang2021-sv,Antonova2023-pk}.
The use of discrete contextualization components requires the implementor of an ASR system to separately train the ASR and contextualization components, to separately optimize their combination, and to take care that all relevant signals, like the input audio, are forwarded from ASR to the contextualization system.

Attention-based biasing~\cite{Pundak2018-fj,Chang2021-zp,Munkhdalai2022-th}, in which the ASR network learns to use inference-time context through attention, and thus requires no separate contextualization components and can be optimized end-to-end by standard backpropagation, has become a popular method for contextualization of end-to-end ASR models.
The core technique has spawned several distinct lines of complementary research.
There are threads of work on improving data efficiency by adapter-style training~\cite{Sathyendra2022-vh} or improving training on synthetic audio derived from text-only data~\cite{Naowarat2023-mj}.
%
Work on precision improvements seeks to lower the rate of over-biasing through hierarchical or gated attention~\cite{Han2022-wv,Munkhdalai2023-ty, wu23e_interspeech,Tong2023-ue,Xu2023-lu,Alexandridis2023-vs}
or through slot triggering~\cite{Lu2023-od,Tong2023-qm};
often, such techniques can improve not only quality but also per-step inference run time~\cite{Munkhdalai2023-ty, Yang2023-nw}.
Notably, Tong et al~\cite{Tong2023-ue} improve WER through an auxiliary slot-level cross-entropy loss.
We do not use slots for selecting context to bias, as we have found it possible to get high performance with hierarchical attention alone; however, in this work, we do apply a phrase-level, rather than slot-level, cross-entropy loss during training to improve WER.
Further we extend the technique to a novel wordpiece-level cross-entropy loss.
Further work aims to improve quality on biased utterances by providing the context encoder with more information than just the graphemic biasing phrases, including phoneme-level features~\cite{Bruguier2019-lh,Hu2019-bt,Chen2019-gp,Pandey2023-wh}, and augmenting the context encoder with semantics-aware embeddings from a pretrained BERT model~\cite{Fu2023-qv}.
Attention-based biasing can also be complemented by shallow fusion with (contextualized) language models~\cite{Xu2023-gp} or combinations of language models and contextualized rescoring~\cite{Dingliwal2023-ke} for further quality improvements.

We distinguish the current work from most of the above by focusing on the inference run time of the \emph{context encoder} which, as noted in Section~\ref{sec:introduction}, is critical to the usability of contextualized ASR.

\subsection{Dual-mode NAM}

\vspace{-0.1in}
\begin{figure}[H]
\begin{minipage}[b]{1.0\linewidth}
  \centering
  \centerline{\includegraphics[scale=0.46]{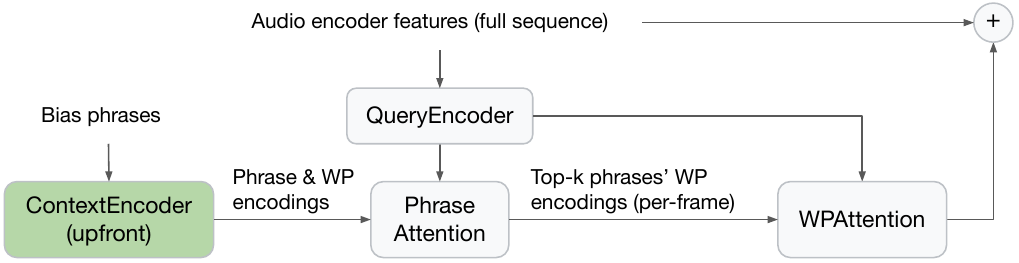}}
\end{minipage}
\caption{Inference with Dual-mode NAM.}
\label{fig:dual_mode_nam}
\end{figure}
\vspace{-0.1in}

Our work builds on Dual-mode NAM (Neural Associative Memory)~\cite{wu23e_interspeech} and its shared query encoder variant from Section 3.4.2 of~\cite{csc_llm}, which we use as a baseline.
Dual-mode NAM computes both phrase encodings $E^p$ and wordpiece (WP) encodings $E^w$ through a fine-grained context encoder, where $E^p$ corresponds to the $cls$ embedding and $E^w$ corresponds to the bias phrase WP embeddings of $Z$; $Z=\{{cls; w_{n, 1}, ..., w_{n, L} }\}^N_{n=1}$, $N$ is the number of bias phrases associated with the utterance and $L$ is the number of WPs per bias phrase.
\begin{eqnarray}
    E^p, E^w = \namoperator{ContextEncoder}(Z)
\end{eqnarray}

The phrase and WP attentions are trained via sampling: The phrase and WP attention contexts $(c^p$, $c^w)$ are added to the audio encoder features $x$ with a probability of $p$ and $1 - p$, respectively.
\begin{eqnarray}
    \label{eq:sampling}
    x^{biased} = x + \namoperator{BernoulliTrial}(c^p, c^w, p)
\end{eqnarray}

During inference (Figure~\ref{fig:dual_mode_nam}), the model leverages the phrase-level attention logits to select per-frame top-$k$ ($k^p$) phrases and feed their WP encodings to the WP attention, where $q_{t,h}^p$ and $K_{t,h}^{p}$ correspond to the projected audio query and $E^p$ encodings.
\begin{eqnarray}
    \label{eq:per_frame_topk}
    I_{t}^p = \namoperator{TopK}(\frac{1}{H} \sum_{h=1}^{H} q_{t,h}^p K_{t,h}^{p}, k^p)
\end{eqnarray}

\section{Methods}

\begin{figure}[H]
\begin{minipage}[b]{1.0\linewidth}
  \centering
  \centerline{\includegraphics[scale=0.46]{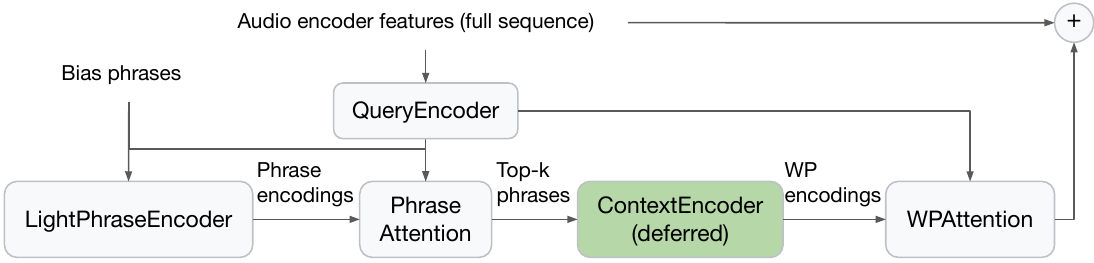}}
\end{minipage}
\caption{Inference with Deferred NAM.}
\label{fig:defered_nam}
\end{figure}

Unlike conventional approaches that uniformly encode all bias phrases beforehand, Deferred NAM (Figure~\ref{fig:defered_nam}) utilizes a lightweight phrase encoder and retrieval process to select the top-k relevant phrases, before invoking the fine-grained context encoder and WP attention at inference. Additionally, Deferred NAM employs cross-entropy losses with its phrase and WP attentions, further boosting WER performance. This design achieves both minimal latency as well as excellent recognition quality.

\vspace{-0.1in}
\vspace{-0.1in}

\subsection{Lightweight phrase encoder}

We adopted~\citet{iyyer-etal-2015-deep}'s  Deep Averaging Network (DAN) as a lightweight encoder to produce phrase encodings $E^p \in \mathbb{R}^{N \times d}$, where the WP embeddings $W=\{{w_{n, 1}, ..., w_{n, L} }\}^N_{n=1}$ are averaged over the $L$ axis and then encoded by a feed-forward network with TANH activation, and $d$ is the WP embedding/encoding dimension. We applied stop gradient ($\bot$) to prevent $LightPhraseEncoder$ from interfering with $\namoperator{ContextEncoder}$ learning of the WP embeddings, which resulted in faster biasing WER convergence.
\begin{eqnarray}
    E^p = \namoperator{LightPhraseEncoder}(\bot(W))
\end{eqnarray}

\subsection{Cross-entropy guided phrase attention}

\begin{algorithm}
\caption{NO\_BIAS-augmented multi-head attention logits with mean-max pooling: $f(q, k)$}
\label{alg:multiheaded_attn}

\textbf{Inputs:} Query $q \in \mathbb{R}^{T\times d_q}$, key $k \in \mathbb{R}^{S \times d_k}$ \\
\textbf{Outputs:} Mean-max pooled logits $z \in \mathbb{R}^{(1+S)}$
\begin{eqnarray}
q_{t,h}^{'} = q_t \Theta_h^Q,  k_{h}^{'} = [\Theta_h^{NB}; k \Theta_h^K] \\
\label{eqn:z_th}
z_{t} = \frac{1}{H} \sum_{h=1}^{H} \frac{q_{t,h}^{'} (k_h^{'})^\top}{\sqrt{d_h}} \\
z = \max_{t=1}^{T} z_{t}
\end{eqnarray}
The trainable parameters are $\Theta_h^Q \in \mathbb{R}^{d_q\times d_h}$, $\Theta_h^K \in \mathbb{R}^{d_k \times d_h}$, $\Theta_h^{NB} \in \mathbb{R}^{d_h}$ (NO\_BIAS token), where $t \in [1..T]$ denotes the time index and $h \in [1..H]$ denotes the attention head index.

\end{algorithm}

As discussed in Section 5 of ~\cite{wu23e_interspeech}, one limitation is that the phrase/WP attentions are trained on fewer examples due to sampling (equation~\ref{eq:sampling}). Another limitation is that the retrieval capability (equation~\ref{eq:per_frame_topk}) is indirectly learned by the ASR loss. We address such limitations by learning the retrieval capability with an explicit loss.

In phrase attention, the relevance between audio query $x^q$ and bias phrases $E^p$ is computed via Algorithm~\ref{alg:multiheaded_attn}. The mean-max pooled logits are then used to compute the softmax cross-entropy loss: 
\begin{eqnarray}
 z^{p} & = & f(x^q, E^p)  \\
    \mathcal{L}_{p} & = & L\_SCE(z^{p}, labels)
\end{eqnarray}
where $labels \in \mathbb{R}^{1 + N}$ corresponds to a probability distribution of the bias labels, with the leading ``1'' being the NO\_BIAS token. During training, a bias phrase is marked as a correct label if it's a longest substring of the transcript truth; the NO\_BIAS token is marked as the correct label if none of the bias phrases is a substring of the transcript truth.

At inference, the global top $k^p$ phrases are used to invoke the context encoder and WP attention.
\begin{equation}
    I^{p}_{global} = TopK(z^{p}_{[2:]}, k^p)
\end{equation}

\subsection{Deferred context encoder}
\label{sec:second_pass}
By offloading phrase encoding to a dedicated lightweight encoder, the context encoder can exclusively focus on generating fine-grained WP encodings for utilization by the WP attention. While the context encoder is trained on the same set of bias phrases $W$ as the phrase encoder, during inference, only the top-$k$ phrases identified through the phrase attention mechanism require encoding.
\begin{equation}
    E^w = \namoperator{ContextEncoder}(W)
\end{equation}
Where $E^w = \{e_{n,1},...,e_{n,L}\}_{n=1}^{N} \in \mathbb{R}^{N \times L \times d}$ represents WP encodings, such that $e_{i,j} \in \mathbb{R}^d$ represents the $i^{\text{th}}$ phrase candidate's wordpiece encoding at position $j \in \{1,...,L\}$.

\subsection{Cross-entropy guided WP attention}

After that, the standard NAM WP attention biasing context $c^w$ is computed and added to the acoustic encoder feature for contextualization.
\begin{eqnarray}
    c^w & = & \namoperator{WPAttention}(x^q, E^w) \\
    x^{biased} & = & x + c^w
\end{eqnarray}
We further augment the $\namoperator{WPAttention}$ with a cross-entropy training loss to boost the likelihood scores of the relevant WPs. Similar to the phrase attention, we first compute the mean-max pooled logits for the WP tokens using Algorithm~\ref{alg:multiheaded_attn}:
\begin{eqnarray}
    z^{w} = f(x^q, E^w) \in \mathbb{R}^{1 + NL} 
\end{eqnarray}
Secondly, the per-phrase average WP logits $\overline{z^{w}_{[2:]}}$ are computed, i.e., by summing the logits of each phrase and then dividing by the phrase's effective sequence length, ignoring padding tokens.
\begin{eqnarray}
\overline{z^{w}_{[2:]}} = \namoperator{PerPhraseAvg}(z^{w}_{[2:]}) \in \mathbb{R}^{N}
\end{eqnarray}
Thirdly, the NO\_BIAS logit $z^{w}_{[1:2]}$ is concatenated with the per-phrase average WP logits $\overline{z^{w}_{[2:]}}$ to form $\overline{z^{w}}$ for calculating the cross-entropy loss.
\begin{eqnarray}
\overline{z^{w}} = [z^{w}_{[1:2]}; \overline{z^{w}_{[2:]}}] \in \mathbb{R}^{1 + N}  \\
    \mathcal{L}_{w} = L\_SCE(\overline{z^{w}}, labels)
\end{eqnarray}
Finally, the total loss is a weighted sum of the ASR, phrase- and WP-level cross-entropy losses.
\begin{eqnarray}
    \mathcal{L}_{total}  = \mathcal{L}_{asr} + \lambda_{p} \mathcal{L}_{p} + \lambda_{w} \mathcal{L}_{w}
\end{eqnarray}

\section{Experiment setup}
\label{sec:experiments}

\subsection{Data sets}


All data sets used aligned with the Privacy Principles and AI Principles in~\cite{privacyprinciples, aiprinciples}. The ASR training data contains 520M anonymized English voice search utterances, totaling 490K hours of speech with an average of 3.4 seconds per utterance. A small percentage of the utterances are human-transcribed and the rest are machine-transcribed by a teacher ASR model~\cite{hwang22c_interspeech}. We evaluate our system on the same multi-context biasing corpora described in Section~3.2.2 of~\cite{Munkhdalai2023-ty}. The corpora consist of three sets: \textbf{WO\_PREFIX}: 1.3K utterances matching prefix-less patterns from \$APPS, \$CONTACTS, and \$SONGS categories (denoted ACS). \textbf{W\_PREFIX}: 2.6K utterances matching prefixed patterns such as ``open \$APPS'', ``call \$CONTACTS'', ``play \$SONGS''. \textbf{ANTI}: 1K utterances simulating general voice assistant queries. Each utterance is assigned up to 3K ACS bias entities. The WO\_PREFIX and W\_PREFIX sets measure in-context performance, where one of the entities appears in the transcript truth; the ANTI set measures anti-context performance, where the utterances are assigned distractor entities only.

\subsection{Model architecture}

Our RNN-T ASR encoder architecture mimics that of Google's Universal Speech Model (USM)~\cite{zhang2023google}.
We use the 128-dimensional log Mel-filterbank energies (extracted from 32ms window and 10ms shift) as the frontend features, which are fed to two 2D-convolution layers, each with strides $(2, 2)$; the resulting feature sequence becomes the input to a stack of 16 Conformer layers~\cite{gulati2020conformer}.
Each conformer layer has 8 attention heads with a total dimension of 1536, and the intermediate dimension of the FFNs is 4 times the attention dimension, yielding a total of 870M parameters in the encoder. The Conformer blocks use local self-attention with a large attention span, and the encoder output has a large enough receptive field to cover the entire utterance. We apply funnel pooling~\cite{dai2020funnel} at the 5th to 7th conformer layers, each with a reduction rate of 2. As a result, the encoder output sequence has a low frame rate of 320ms. The model uses a $|V|^2$ embedding decoder~\cite{Rami21}, i.e., the prediction network computes LM features based on two previous non-blank tokens.  The output vocabulary size consists of 4096 lowercase wordpieces.

\vspace{-0.1in}
\begin{figure}[H]
\begin{minipage}[b]{1.0\linewidth}
  \centering
  \centerline{\includegraphics[scale=0.55]{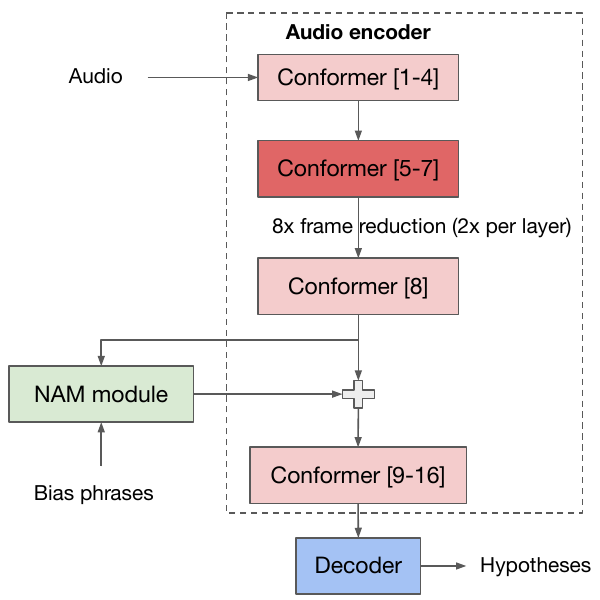}}
\end{minipage}
\caption{Non-streaming RNN-T + NAM module. }
\label{fig:nam_module}
\end{figure}
\vspace{-0.2in}

Both baseline and experiment NAM modules are placed in between 8th and 9th Conformer layer as shown in Fig.~\ref{fig:nam_module}. In accordance with~\cite{wu23e_interspeech}, bias phrases are sampled from reference transcripts during training; the bias strength $\lambda = 0.6$ is introduced at inference: $x^{biased} = x + \lambda c^w$. 

The models are developed using the open-source Lingvo toolkit~\cite{shen2019lingvo}, trained on 16x16 cloud TPU v3~\cite{tpu} with a global batch size of 4096 for 240K steps (3 days). All parameters are randomly initialized and optimized by the Adafactor optimizer~\cite{adafactor18}.

\subsection{Baseline: Dual-mode NAM}
\label{sec:dual_nam_module}

We configured dual-mode NAM B1, B2 modules to be the WER, latency baseline of Deferred NAM. 

The B1 module has 100.8M parameters, with $\namoperator{QueryEncoder}$ (89.8M): a 2L Conformer with a model dimension of 1536, hidden dimensions of [6144, 3072] for the internal feed-forward layers; $\namoperator{ContextEncoder}$ (4M): 3L Conformer (3M) with a model, hidden dimension of 256, 512 and the WP embedding table (1M) with a vocab size of 4096. $\namoperator{PhraseAttention}$/$\namoperator{WPAttention}$ (5.5M each): 8 heads with a per-head dimension of 192. 

The B2 module is similar to B1 except that $\namoperator{ContextEncoder}$ contains 1L Conformer (1M) instead of 3L. In Table~\ref{tbl:dual_nam_wers}, we show that reducing the context encoder from 3L to 1L for a premature latency optimization would negatively impact the in-context WERs by up to 26.8\% relative ($3.0 \rightarrow 4.1$).

\vspace{-0.1in}
\begin{table}[H]
\centering
\begin{tabular}{l|rrr}
\toprule
Expt & \textbf{B1} & B2 \\
\midrule
ANTI & 2.3 &  1.9   \\
WO\_PREFIX & \textbf{3.0} &  4.1 \\
W\_PREFIX & \textbf{2.4} &  2.9   \\
\bottomrule
\end{tabular}
\caption{Dual-mode NAM average WERs at $k^p$=32, computed by averaging over scenarios where (150, 300, 600, 1.5K, 3K) bias entities are provided per utterance.}
\vspace{-0.1in}
\label{tbl:dual_nam_wers}
\end{table}

\subsection{Experiment: Deferred NAM}
\label{sec:deferred_nam_mdule}

We explore variants D1--D3 to study the impact of each proposed training loss. When compared to B1, Deferred NAM adds a $\namoperator{LightPhraseEncoder}$: a 4L DAN phrase encoder (788K parameters) with TANH activation, a model, hidden dimension of 256; the $\namoperator{PhraseAttention}$ (2.8M parameters) contains only parameters as described in Algorithm~\ref{alg:multiheaded_attn}; the $\namoperator{ContextEncoder}$ consists of only a 1L Conformer (1M) identical to B2, which we found is sufficient to outperform the WERs of B1 (3L).

\begin{itemize}
    \item[\textbf{D1}] The model learns retrieval capability via equation~\ref{eq:sampling}, where $p=0.3$ and $\mathcal{L}_{total} = \mathcal{L}_{asr}$.
    \item[\textbf{D2}] The model learns retrieval capability through cross-entropy guided phrase attention (CE-PA): $\mathcal{L}_{total}  = \mathcal{L}_{asr} + 0.1 \mathcal{L}_{p}$.
    \item[\textbf{D3}] D2 with cross-entropy guided WP attention (CE-WA): $\mathcal{L}_{total}  = \mathcal{L}_{asr} + 0.1 \mathcal{L}_{p} +  0.1 \mathcal{L}_{w}$.
\end{itemize}

\section{Results}

\subsection{Quality}
\label{sec:quality}

As shown in Table~\ref{tbl:deferred_nam_wer_summary}, the base Deferred NAM~(D1) already outperforms the best Dual-mode NAM~(B1)'s average WERs by up to 20.8\% relative ($2.4 \rightarrow 1.9$), while using fewer parameters in total. By augmenting the model with the CE-PA loss, D2 improves over D1 by up to 11.5\% relative ($2.6 \rightarrow 2.3)$. With addition of CE-WA loss, D3 improves over D2 by 16.7\% relative ($1.8 \rightarrow 1.5$). As a result, the best Deferred NAM~(D3) improves over the best Dual-mode NAM~(B1) by up to 37.5\% relative ($2.4 \rightarrow 1.5$) on in-context recognition, and 21.7\% on anti-context recognition ($2.3 \rightarrow 1.8$).

\vspace{-0.1in}
\begin{table}[H]
\centering
\begin{tabular}{l|rrrrr}
\toprule
Expt  &  B1 & D1 & D2 & D3  \\
\midrule
ANTI &  2.3 &  1.9 &  \textbf{1.8} &  \textbf{1.8}  \\
WO\_PREFIX  & 3.0 &  2.6 &  2.3 &  \textbf{2.0} \\
W\_PREFIX  & 2.4 &  1.9 &  1.8 &  \textbf{1.5} \\
\bottomrule
\end{tabular}
\caption{Average WERs (\# entities = 0 excluded) at $k^p$=32 comparing Dual-mode (B1) and Deferred NAM (D1--D3). WER breakdown is shown in Table~\ref{tbl:wer_breakdown}.}
\vspace{-0.2in}
\label{tbl:deferred_nam_wer_summary}
\end{table}

\begin{table}[H]
\centering
\begin{tabular}{lr|r|rrr}
\toprule
Expt & \# & B1 &   D1 & D2 & D3  \\
\midrule
\multirow{6}{*}{ANTI}         
            & 0    &   \textbf{1.4} &   \textbf{1.4} &   1.5 &   1.6 \\
            & 150  &   1.7 &   1.6 &   \textbf{1.5} &   1.7 \\
            & 300  &   2.0 &   \textbf{1.6} &   \textbf{1.6} &   1.8 \\
            & 600  &   2.4 &   2.0 &   \textbf{1.7} &   1.9 \\
            & 1.5K &   2.6 &   \textbf{1.9} &   \textbf{1.9} &   \textbf{1.9} \\
            & 3K &   2.9 &   2.2 &   2.3 &   \textbf{1.9} \\
\midrule
        & 0    &  \textbf{21.1} &  21.6 &  \textbf{21.1} &  21.3 \\
\multirow{3}{*}{WO\_}   & 150  &   1.8 &   2.0 &   1.8 &   \textbf{1.6} \\
\multirow{3}{*}{PREFIX} & 300  &   2.0 &   2.0 &   2.0 &   \textbf{1.7} \\
        & 600  &   2.7 &   2.3 &   2.3 &   \textbf{1.8} \\
        & 1.5K &   3.4 &   3.1 &   2.4 &   \textbf{2.3} \\
        & 3K &   4.9 &   3.5 &   3.1 &   \textbf{2.7} \\
\midrule
        & 0    &   9.8 &   9.9 &  10.0 &   9.9 \\
\multirow{3}{*}{W\_}        & 150  &   1.5 &   1.6 &   1.5 &   \textbf{1.3} \\
 \multirow{3}{*}{PREFIX}        & 300  &   1.8 &   1.6 &   1.6 &   \textbf{1.4} \\
    & 600  &   2.1 &   1.8 &   1.7 &   \textbf{1.5} \\
       & 1.5K &   2.8 &   2.1 &   1.8 &   \textbf{1.6} \\
       & 3K &   3.6 &   2.3 &   2.2 &   \textbf{1.9} \\
\bottomrule
\end{tabular}
\caption{Detailed WER breakdown at $k^p$=32 on the best Dual-mode (B1) and Deferred NAM (D1--D3), where each utterance is assigned up to 3K bias entities.}
\vspace{-0.1in}
\label{tbl:wer_breakdown}
\end{table}

We also show Deferred NAM's 1st pass retrieval recall performance in Table~\ref{tbl:retrieval_recall}. Interestingly, although there are sizable retrieval performance increases at smaller $k^p \leq 5$ from D1 to D2 (up to 25.6\% relative, e.g., $69.2 \rightarrow 86.9$), D1's recall performance at $k^p = 32$ (where the WERs are evaluated at) is already quite high and left little room for further improvement (up to 2.3\% relative, e.g., $97.4 \rightarrow 99.6$). On the other hand, D3's retrieval performance is more in line with D2, as expected.

Given the slightly-improved or similar recall performance at $k^p = 32$, we attribute D2's WER improvement over D1 to better-regulated audio/text embeddings (due to CE-PA) and increased data exposure of the 2nd-pass contextualization, i.e., D2 has full training data exposure while D1's phrase and WP attentions are only trained 30\% and 70\% of the time, respectively, due to sampling. D3's WER improvement over D2 is more obvious as the CE-WA loss is directly applied to the WP attention. 

\vspace{-0.1in}
\begin{table}[H]
\centering
\begin{tabular}{lr|rrr}
\toprule
Expt & $k^p$ &   D1 & D2 & D3  \\
\midrule
           & 1    & 73.0  & 91.5  & \textbf{93.4}  \\
WO\_PREFIX & 5   &  92.8 & 98.7  & \textbf{98.9}  \\
           & 32   &  98.9 & \textbf{99.7}  &  \textbf{99.7} \\

\midrule
          & 1     &  69.2  & 86.9  & \textbf{87.9}  \\
W\_PREFIX & 5   &  89.8 &  \textbf{98.0}  &  97.7  \\
          & 32   &   97.4 & \textbf{99.6}   &  99.5  \\

\bottomrule
\end{tabular}
\caption{Retrieval recall performance of D1--D3 by $k^p$ for in-context test-sets at 3K bias entities per utterance.}
\vspace{-0.1in}
\label{tbl:retrieval_recall}
\end{table}

\subsection{Inference latency}

\vspace{-0.1in}
\begin{table}[H]
\centering
\begin{tabular}{l|rr}
\toprule
& \multicolumn{2}{c}{\# phrases} \\
Deferred NAM latency (ms) & 3K  & 20K\\
\midrule
$\namoperator{QueryEncoder}$ & 2.3 & 2.3 \\
$\namoperator{LightPhraseEncoder}$  & 3.5 & 22.8 \\
$\namoperator{PhraseAttention}$  & 0.9 & 5.2 \\
$\namoperator{ContextEncoder}$ & 1.3 & 1.3\\
$\namoperator{WPAttention}$ & 0.7 & 0.7 \\
\midrule
Total & 8.7 & 32.3\\
\bottomrule
\end{tabular}
\caption{Latency of Deferred NAM (D3), at $k^p$=32.}
\vspace{-0.1in}
\label{tbl:latency}
\end{table}

The ASR is benchmarked on 1x1 cloud TPU V3 at bfloat16, with batch size (number of utterances) 8; each utterance has 512 time steps (15.36s of audio), and each bias phrase has a length of 16 WPs. As shown in Table~\ref{tbl:latency}, the total latency of Deferred NAM at processing 3K and 20K bias phrases is only 8.7ms and 32.3ms, respectively. On the other hand, encoding all bias phrases up front (Dual-mode NAM) is significantly slower, i.e., B1's 3L Conformer context encoder latency alone is 214ms and 1549ms; B2's 1L Conformer context encoder latency alone is at 72ms and 520ms. Overall, Deferred NAM provides a speedup of at least 8.3X and 16.1X over Dual-mode NAM's best-case latency scenario (B2), and surpasses the best-case quality scenario (B1), as discussed in Section~\ref{sec:quality}.

\subsection{Exploration: NO\_BIAS filter}
\label{nobias_filter}

We investigated gating the WP attention with the phrase-level NO\_BIAS token, where a bias phrase is deemed ``active'' if its per-frame logit is higher than that of the NO\_BIAS token ($z^p_{t}[1:2]$) at any frame. Method 1 (M1) filters inactive phrases before WP attention~(equation~\ref{eq:no-bias-filter:method-1}). However, with fewer WPs to attend to, the remaining ones receive higher attention probabilities.
This could degrade anti-context WERs if the accuracy of the NO\_BIAS token is not sufficiently high, i.e., under condition $m$, only 43.6\% of utterances are deemed inactive for D3 at ANTI (3K). Therefore, we explored Method 2 (M2), which zeroes out the WP attention \emph{values} of inactive phrases, leaving the probabilities of the WP \emph{keys} unaffected~(equation~\ref{eq:no-bias-filter:method-2}).
\begin{eqnarray}
    m = \vee_{t=1}^T (z^p_{t}[2:] > z^p_{t}[1:2]) \in \mathbb{R}^{N} \\
   Method~1: (I^p_{global})' = Filter(I^p_{global}, m) \label{eq:no-bias-filter:method-1} \\
   Method~2: (v^w)' = v^w_{i,j} \Theta^V \circ m_i \label{eq:no-bias-filter:method-2}
\end{eqnarray}
$v^w_{i,j}$: The attention value of $i$-th phrase at $j$-th WP; $\Theta^V$: The value projection matrix of WP attention.  

\vspace{-0.1in}
\begin{table}[H]
\centering
\begin{tabular}{l|rr|rrr}
\toprule
Expt  &  \multicolumn{2}{c}{D1} & \multicolumn{3}{c}{D3}  \\
      &  N/A & M1 & N/A & M1 & M2  \\
\midrule
ANTI &  1.9  &   \textbf{1.4}   &   1.8    &  1.9    & 1.7   \\
WO\_PREFIX  &  2.6     &   12.9    &   2.0  &  \textbf{1.8}  &   2.1      \\
W\_PREFIX  &   1.9     &  8.3    &    \textbf{1.5} &   \textbf{1.5}    &   1.6    \\
\bottomrule
\end{tabular}
\caption{Average WERs (\# entities = 0 excluded) for Deferred NAM (D1 \& D3) with NO\_BIAS filtering.}
\vspace{-0.1in}
\label{tbl:nobias_wers}
\end{table}

Table~\ref{tbl:nobias_wers} shows Deferred NAM's (D3) ability to use the phrase-level NO\_BIAS token in inference. D3-M1 shows a relative improvement of up to 10\% ($2 \rightarrow 1.8$) in in-context WERs, with a relative anti-context decline of 5.6\% ($1.8 \rightarrow 1.9$). Conversely, D3-M2 shows a relative improvement of 5.6\% ($1.8 \rightarrow 1.7$) in anti-context WERs with a relative increase of up to 6.7\% ($1.5 \rightarrow 1.6$) in in-context WERs. Notably, the NO\_BIAS token's logit dominates in D1 (similar to Dual-mode NAM~\cite{wu23e_interspeech}) due to the lack of supervised training (i.e., CE-PA), leading to a substantial relative increase of up to 396.2\% ($2.6 \rightarrow 12.9$) in in-context WERs.
\section{Conclusion}
\label{sec:conclusion}

We proposed a low-latency attention-based contextual ASR system, augmented with phrase- and WP-level cross-entropy losses, which can handle thousands of bias phrases within milliseconds while achieving up to 37.5\% relative average WER reduction. This demonstrates the potential for enhancing ASR in real-world applications requiring fast and accurate contextual speech recognition.

\bibliography{anthology,custom}




\end{document}